\documentclass[conference]{IEEEtran}
\IEEEoverridecommandlockouts

\usepackage{cite}
\usepackage{amsmath,amssymb,amsfonts}
\usepackage{algorithmic}
\usepackage{graphicx}
\usepackage{textcomp}
\usepackage[export]{adjustbox} 
\usepackage{xcolor}
\usepackage{float}
\usepackage{hyperref}
\usepackage[export]{adjustbox}
\usepackage{subcaption}
\def\BibTeX{{\rm B\kern-.05em{\sc i\kern-.025em b}\kern-.08em
    T\kern-.1667em\lower.7ex\hbox{E}\kern-.125emX}}
\begin{document}

\title{Synthesising Handwritten Music with GANs: A Comprehensive Evaluation of CycleWGAN, ProGAN, and DCGAN\\

}

\author{\IEEEauthorblockN{1\textsuperscript{st} Elona Shatri}
\IEEEauthorblockA{\textit{Centre for Digital Music} \\
\textit{Queen Mary University of London}\\
London, UK \\
e.shatri@qmul.ac.uk}
\and
\IEEEauthorblockN{2\textsuperscript{nd} Kalikidhar Reddy Palavala}
\IEEEauthorblockA{\textit{Big Data Science} \\
\textit{Queen Mary University of London}\\
London, UK \\
k.palavala@se23.qmul.ac.uk}
\and
\IEEEauthorblockN{3\textsuperscript{rd} Gy\"orgy Fazekas}
\IEEEauthorblockA{\textit{Centre for Digital Music} \\
\textit{Queen Mary University of London}\\
London, UK \\
george.fazekas@qmul.ac.uk}
}


\maketitle

\begin{abstract}

The generation of handwritten music sheets is a crucial step toward enhancing Optical Music Recognition (OMR) systems, which rely on large and diverse datasets for optimal performance. However, handwritten music sheets, often found in archives, present challenges for digitisation due to their fragility, varied handwriting styles, and image quality. This paper addresses the data scarcity problem by applying Generative Adversarial Networks (GANs) to synthesise realistic handwritten music sheets. We provide a comprehensive evaluation of three GAN models—DCGAN, ProGAN, and CycleWGAN—comparing their ability to generate diverse and high-quality handwritten music images. The proposed CycleWGAN model, which enhances style transfer and training stability, significantly outperforms DCGAN and ProGAN in both qualitative and quantitative evaluations. CycleWGAN achieves superior performance, with an FID score of 41.87, an IS of 2.29, and a KID of 0.05, making it a promising solution for improving OMR systems.
\end{abstract}

\begin{IEEEkeywords}
Image Translation, Generative Adversarial Networks, Sheet Music.
\end{IEEEkeywords}

\section{Introduction}\label{sec:Introduction}

Handwritten music has been a critical medium for composing and preserving musical works, with roots tracing back to Babylonia in 1200 BC. In today’s digital age, converting these handwritten scores into digital formats is imperative for ensuring their accessibility and long-term preservation. However, the manual transcription of such scores is both costly and time-consuming, driving a significant demand for automated solutions \cite{b5}.

Optical Music Recognition (OMR) systems play a crucial role in transforming scanned music scores into structured digital formats, such as MIDI\footnote{MIDI: \url{https://en.wikipedia.org/wiki/MIDI}}, MusicXML\footnote{MusicXML: \url{https://www.musicxml.com/}}, and MEI\footnote{MEI: \url{https://music-encoding.org/}} \cite{b19}, enabling compositions to be searchable, editable, and easily shared. While recent advancements in deep learning have led to significant improvements in OMR systems for printed scores \cite{b18}, \cite{b38}, the recognition of handwritten music remains a considerably under-researched area. This limitation is primarily due to the scarcity of annotated handwritten music datasets and challenges such as varying line thickness, curvature, and quality \cite{b5}. This data limitation hampers the performance of OMR systems, which require diverse and high-quality training data to generalise effectively.

The motivation behind this work stems from the need to bridge this gap and support OMR systems in achieving similar success with handwritten scores as they have with printed ones. The scarcity of annotated datasets is a fundamental barrier, and creating diverse, realistic synthetic handwritten music data can significantly aid in training OMR models. Generating such data, however, presents unique challenges. For instance, while one could consider generating MIDI files and overlaying handwritten patterns, this approach faces two significant obstacles: the absence of a reliable pipeline transitioning from MIDI to MusicXML or MEI for producing handwritten scores, and the need for a dataset encompassing diverse handwriting styles to bolster model robustness. This study addresses these challenges by including a variety of styles from the MUSCIMA dataset.

To meet these demands, Generative Adversarial Networks (GANs) offer a promising solution by synthesising high-quality, realistic handwritten music images that augment existing datasets. These synthetic images not only enhance the training of OMR systems but also find applications in fields such as graphic design and publishing, where specific handwritten music styles may be desired for artistic or editorial purposes. This paper explores advanced GAN architectures—specifically DCGAN, ProGAN, and CycleWasserstein GAN (CycleWGAN)—to improve the generation of diverse handwritten music images suitable for OMR system training.

While these GAN architectures have been proposed in other works, adapting them effectively to the music domain presents significant challenges. Handwritten music data, being limited and difficult to procure, requires substantial preprocessing. We addressed these issues by modifying the CycleGAN architecture to incorporate Wasserstein adversarial loss functions, instance normalisation, and multiple residual blocks in the generator. These modifications improved training stability and style transfer quality, achieving better results than CycleGAN variants described in the literature.

The primary contribution of this study lies in applying CycleWGAN, a CycleGAN variant with the Wasserstein distance, to handwritten music generation. This variant ensures enhanced training stability and higher-quality style transfer. Unlike previous research focused solely on individual GAN models, we provide a comparative analysis of three GAN architectures—DCGAN, ProGAN, and CycleWGAN—to evaluate their performance in generating realistic, diverse handwritten music sheets.

Our findings expand the current handwritten music datasets by producing more realistic and varied synthetic samples. This study offers valuable insights into how different GAN models perform in this domain, providing practical guidance for future development of OMR systems.

\section{Background}

OMR systems, such as PhotoScore\footnote{PhotoScore: \url{https://www.neuratron.com/photoscore.htm}} and PlayScore\footnote{PlayScore: \url{https://www.playscore.co/}}, have shown significant success in recognising printed music. However, these systems struggle substantially with handwritten scores due to the inherent variability in handwriting styles and the complexity of musical notations \cite{b20}. The challenges of irregular spacing, varying line thickness, and symbol curvature in handwritten scores require a more flexible approach than those employed in printed music recognition. 

While recent advances in deep learning have improved OMR performance for printed scores, applying these techniques to handwritten scores has remained limited, largely due to the scarcity of annotated datasets. Handwritten scores, particularly historical ones, often vary widely in terms of notation style, making it difficult to train models that generalise well across different handwriting types.

Deep learning techniques, including Bidirectional Long Short-Term Memory (BLSTM) networks \cite{b8}, sequence-to-sequence models \cite{b39}, object detection and instance segmentation \cite{b45} and transformers \cite{b38}, have significantly enhanced OMR for printed music. However, these methods have not yet been adapted effectively for handwritten scores due to the lack of large, diverse datasets that can capture the full range of handwriting styles and musical symbols. Early attempts to apply convolutional and recurrent neural networks, such as the work by \cite{b2}, were restricted by relatively small datasets, like MUSCIMA++ \cite{b1}, which limited their scope to recognising simple staves and notation patterns.

To address this data scarcity and variability, GANs have emerged as a promising solution for generating synthetic training data. GANs, introduced by \cite{b37}, are well-suited to this task due to their ability to learn complex data distributions in an unsupervised manner. A GAN consists of two competing networks: the generator, which produces synthetic data, and the discriminator, which distinguishes between real and generated data. This adversarial process drives the generator to create high-quality, realistic images that can be used to augment training datasets.



Several GAN architectures have been proposed to improve the quality of generated images, particularly in cases where datasets are scarce or highly variable. Deep Convolutional GANs (DCGANs) \cite{b36} were among the first to incorporate deep convolutional layers into the traditional GAN framework, improving the spatial coherence of generated images. This makes DCGAN a logical starting point for generating handwritten music, where maintaining the spatial structure of musical symbols is critical. DCGANs have been successfully applied to generate handwritten text in resource-scarce languages like Urdu \cite{b17} and Arabic \cite{b42}, showing potential in domains where training data is limited. However, DCGAN's relatively simple architecture may struggle to capture the complexities of handwritten music notation, particularly in cases involving more intricate symbols and varied handwriting styles.

ProGANs \cite{b4} introduced a progressive training approach, starting with low-resolution images and gradually increasing resolution as the model learns. This technique allows for greater training stability and improved image quality over time, making it highly suitable for generating detailed handwritten music images. ProGAN’s progressive resolution strategy has been effective in fields like medical imaging and satellite imagery \cite{b15}, both of which require high-resolution output and face challenges similar to handwritten music recognition, where image detail is critical. By progressively learning to generate more detailed features, ProGAN can produce high-resolution musical symbols that capture the nuances of human handwriting.

Despite these advancements, early approaches to synthesising handwritten music, such as Mashcima \cite{b35}, which combined symbol masks from the MUSCIMA++ dataset \cite{b1} with annotations from PrIMuS \cite{b43}, were constrained by their reliance on predefined symbols. This approach lacked the flexibility required to generate diverse handwriting styles and was unable to capture the variability needed for effective OMR training. More recent work by \cite{b34} using autoencoders for handwritten symbol generation yielded noisy results, highlighting the need for more advanced models capable of generating cleaner, more accurate representations. Recent work has also been done using Enhanced Super-Resolution (ESRGAN) \cite{b46} to generate crops, which are then stitched together to form staves.

CycleGANs \cite{b3}, a class of GANs designed for unpaired image-to-image translation tasks, have been successfully used to translate printed music into handwritten versions \cite{b7}, capturing stylistic elements specific to handwritten scores. While CycleGANs are effective at translating styles, challenges such as hyperparameter tuning, training instability, and imperfect style transfer remain. These limitations motivate the need for more sophisticated GAN architectures, such as the CycleWGAN, which employs Wasserstein loss \cite{b24} for improved training stability and more accurate style transfer.

Given the complexity of handwritten music scores and the challenges posed by data scarcity, GANs represent a powerful approach for generating diverse, high-quality synthetic datasets. 

\section{Methodology}

This study utilises advanced Generative Adversarial Network (GAN) architectures to generate and translate handwritten music images from printed scores. The primary goal is to address the data scarcity problem in OMR by generating diverse and realistic handwritten music images that can be used to train OMR systems. The effectiveness of three different GAN models—DCGAN, ProGAN, and CycleWGAN—is compared in terms of their ability to generate high-quality handwritten music data.

\subsection{Datasets and Data Augmentation}

The CVC-MUSCIMA dataset, which contains a variety of complex handwritten music patterns, is used as the primary source for modelling handwritten music. To perform image-to-image translation between printed and handwritten scores, CVC-MUSCIMA (handwritten) is paired with the DoReMi dataset \cite{b6}, containing 6432 printed music sheet pages. The combination of these datasets allows the models to learn both the structure of printed scores and the stylistic variations of handwritten music.

\begin{figure}[H]
    \centering
    \hspace{0.5cm}
    \subfloat[]{%
        \includegraphics[scale=0.2]{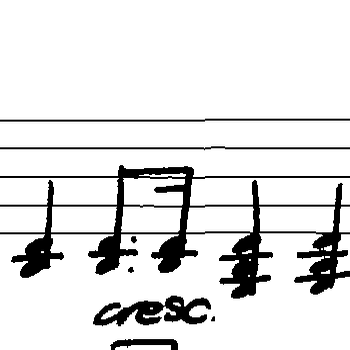}
        \label{hw_score}
        }
    \hspace{0.5cm}
    \subfloat[]{%
        \includegraphics[scale=0.3]{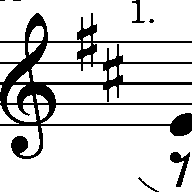}
        \label{printed_score}
    }
    \caption{(a) Handwritten image crop from the CVC-MUSCIMA dataset obtained after the data augmentation process. (b) Printed image crop from the DOREMI dataset obtained after the data augmentation process.}
    \label{fig:comparison}
\end{figure}

To mitigate the challenge of data scarcity, extensive data augmentation techniques are applied. Each rectangular staff image is converted into multiple square crops, increasing the diversity of the training dataset and ensuring uniformity for GAN input. This augmentation process generates 16,826 handwritten and 28,683 printed images. All images are converted from RGB to grayscale to reduce computational complexity and prevent unnecessary transformations, such as flipping or rotating, which could distort the semantic content of the musical symbols. Figures \ref{hw_score} and \ref{printed_score} show sample cropped handwritten and printed images used for training.

\subsection{GAN Models for Handwritten Music Generation}
\subsubsection{DCGAN: Baseline Architecture}

The Deep Convolutional GAN (DCGAN) \cite{b36} was selected as the baseline model due to its established effectiveness in generating realistic images. DCGAN improves upon the original GAN architecture by introducing convolutional layers in both the generator and discriminator networks. These layers are better suited for capturing local dependencies within images, making DCGAN particularly useful for tasks that require spatial coherence, such as generating musical notations where the relative positioning of symbols, staff lines, and notes is crucial.

DCGAN’s generator begins by transforming a random noise vector into a synthetic image through a series of upsampling and convolutional layers, while the discriminator is tasked with distinguishing between real and generated images. The adversarial training process drives the generator to improve the quality of its outputs as the discriminator becomes increasingly adept at identifying fake images.

The adversarial training process is governed by the following loss functions for the discriminator and generator:

\begin{equation}
\mathcal{L}_D = -\mathbb{E}_{x \sim p_{\text{data}}(x)} [\log D(x)] - \mathbb{E}_{z \sim p_z(z)} [\log (1 - D(G(z)))]
\end{equation}

\begin{equation}
\mathcal{L}_G = -\mathbb{E}_{z \sim p_z(z)} [\log D(G(z))]
\end{equation}

Here, \(x \sim p_{\text{data}}(x)\) refers to real data samples drawn from the true data distribution, and \(z \sim p_z(z)\) represents noise vectors sampled from a predefined distribution, typically Gaussian. \(G(z)\) is the generator output, while \(D(x)\) is the discriminator’s estimate of the probability that \(x\) is real. \(\mathbb{E}\) denotes the expectation operator. The discriminator loss \(\mathcal{L}_D\) measures how well the discriminator distinguishes real from generated images, and the generator loss \(\mathcal{L}_G\) reflects how well the generator can fool the discriminator.

Despite its advantages in generating visually coherent images, DCGAN is inherently limited by its architecture when applied to complex tasks such as generating high-resolution handwritten music sheets. Handwritten music often contains intricate details—such as clefs, and dynamic markings—that vary significantly between samples. DCGAN’s fixed resolution output and relatively simple generator structure can lead to issues like mode collapse, where the model generates limited variations of symbols, and low-resolution output, which affects the clarity of finer musical details.

These limitations become particularly problematic for OMR systems, which require high-resolution input to accurately distinguish between similar musical symbols. DCGAN's inability to capture fine-grained details across diverse handwriting styles further reduces its applicability in generating realistic training data for OMR tasks. Therefore, while DCGAN serves as a useful baseline, more advanced architectures like ProGAN and CycleWGAN are necessary to improve both the quality and diversity of generated handwritten music images.

\subsubsection{ProGAN: Progressive Image Generation}

To overcome the limitations of DCGAN, ProGAN \cite{b4} introduces a progressive training approach that allows the model to generate high-resolution images more effectively. Unlike traditional GAN models, which generate images at a fixed resolution, ProGAN starts with low-resolution images (4x4 pixels) and incrementally increases the resolution throughout the training process. This progressive scaling stabilises the model's learning, as it enables the generator and discriminator to focus on coarse-level features initially and gradually refine finer details as the resolution increases. This approach significantly reduces the risk of mode collapse, a common issue in GAN training, where the model generates limited and repetitive outputs.

The ability to incrementally scale the resolution is particularly beneficial for tasks involving handwritten music generation, where the model needs to capture both global structure (e.g., staff lines, overall note placement) and intricate local details (e.g., noteheads, beams, and clefs). In this study, the resolution is scaled from 4x4 to 128x128 pixels, which balances computational efficiency with the need for detailed image generation. This is a departure from the original ProGAN implementation, which extended resolutions to 1024x1024; the reduced resolution here is appropriate for the specific needs of music notation while optimising memory usage.

ProGAN also incorporates several techniques that enhance image quality and stabilise training such as Latent Vector Reduction, where the latent vector size, 'z', is reduced from 512 to 256 dimensions to optimise performance under memory constraints without compromising the generator's ability to produce diverse outputs. Then, Pixel Normalisation, Weight Scaling, and Leaky ReLU Activations are applied within the convolutional layers of the generator to maintain the stability of gradients during training and improve the quality of the generated images. Finally, a fade-in mechanism is employed when increasing the image resolution, ensuring that transitions between resolutions are smooth. Without this, sudden jumps in resolution could destabilise training, leading to abrupt changes in the generated images.

To further enhance training stability, ProGAN uses the Wasserstein GAN with Gradient Penalty (WGAN-GP) framework \cite{b22}. The Wasserstein distance, as opposed to traditional GAN losses, provides a smoother and more informative gradient for the generator. This not only reduces the likelihood of mode collapse but also prevents the discriminator from overpowering the generator, a common issue when training GANs on complex data like handwritten music. The gradient penalty ($\mathcal{L}_{GP}$) enforces Lipschitz continuity, ensuring that the model remains stable even with high-resolution images. The corresponding loss functions for the discriminator and generator are:

\begin{align}
    \mathcal{L}_D &= -\mathbb{E}[\text{D}(\mathbf{x}_{real})] + \mathbb{E}[\text{D}(\mathbf{x}_{fake})] \nonumber \\
    &\quad + \lambda_{GP} \cdot \mathcal{L}_{GP} \\
    \mathcal{L}_G &= -\mathbb{E}[\text{D}(\mathbf{x}_{fake})]
\end{align}

Here, $\mathcal{L}{GP}$ represents the gradient penalty, applied with a coefficient $\lambda{GP} = 10$ \cite{b22}. This stabilisation is crucial for generating realistic handwritten music images, where both symbol accuracy and visual coherence are necessary. By employing these techniques, ProGAN is able to generate higher-quality, more detailed images compared to DCGAN, making it more suitable for complex image generation tasks like handwritten music generation.

\subsubsection{CycleWGAN: Improved Style Transfer with Wasserstein Loss}

\begin{figure*}[h]
    \centering
    \includegraphics[width=0.9\textwidth]{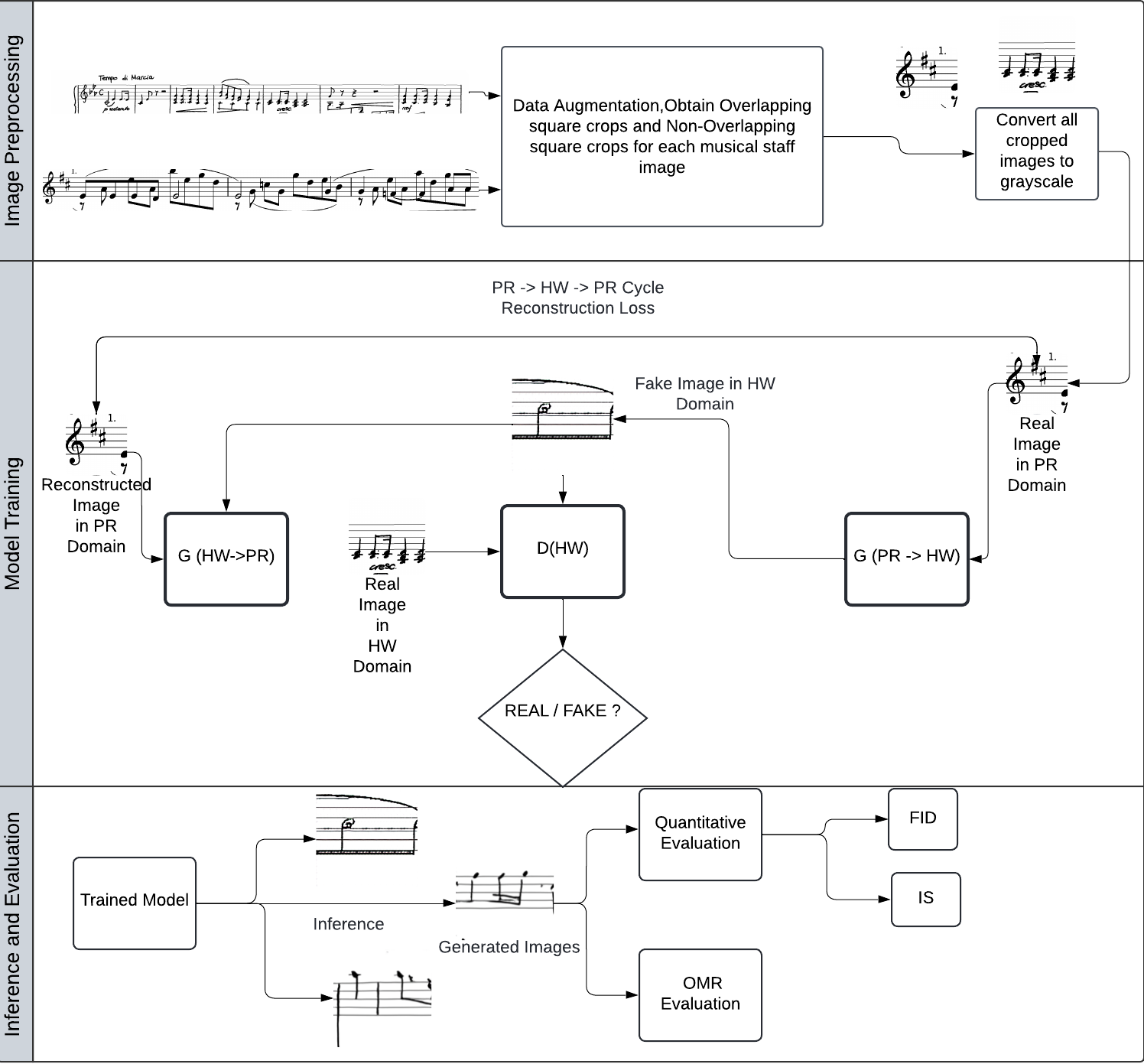}
        \caption{CycleWGAN pipeline architecture used for handwritten and printed scores.}
        \label{fig:cyclegan-arch}

\end{figure*}

\begin{figure}[H]
  \centering
  \includegraphics[width=0.9\linewidth]{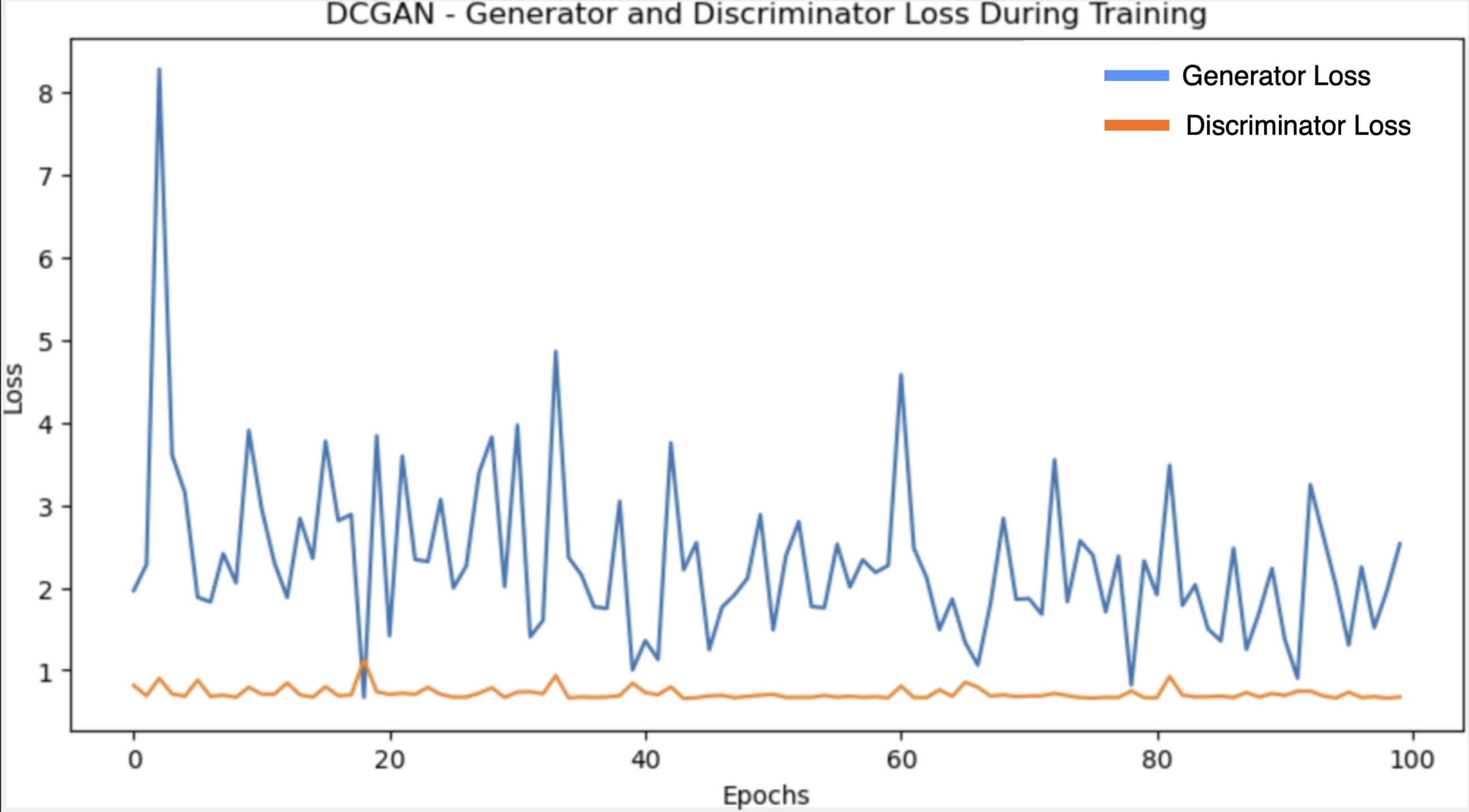}
  \caption{Training loss of DCGAN over 100 epochs.}   
  \label{fig:picture10}
\end{figure}

To further improve the generation of handwritten music images, we employ CycleWGAN, a modified version of CycleGAN \cite{b26}. While CycleGAN is effective for unpaired image-to-image translation, it can suffer from unstable training and mode collapse when dealing with complex image structures, such as musical symbols.

CycleWGAN addresses these issues by replacing the standard adversarial losses with the Wasserstein distance \cite{b24}, which enhances training stability and reduces the likelihood of mode collapse. Additionally, a cycle consistency loss, optimised using the L2 norm (mean squared error (MSE)) \cite{b3}, ensures that the translation between printed and handwritten music preserves key musical details, allowing the model to generate more realistic style transfers.

The CycleWGAN architecture consists of two generators (Gp and Gh) and two discriminators (Dp and Dh), given in Figure \ref{fig:cyclegan-arch}. The generators convert between printed and handwritten styles, while the discriminators distinguish real images from generated ones in both domains. The model also employs a ResNet-based architecture \cite{b32}, incorporating nine residual blocks to better capture the subtle variations in handwriting styles.

The loss functions governing the training process are two adversarial Wasserstein losses  and cycle consistency loss, given as follows:
\vspace{-0.2em}
\begin{align*}
\mathcal{L}_{(G_p, D_p, P, H))} &= \mathbb{E}_{h \sim p_{\text{H}}(h)} [ D_h (h)] - \mathbb{E}_{p \sim p_{\text{P}}(p)} [D_h(G_p(p))] \\
\mathcal{L}_{(G_h, D_h, H, P))} &= \mathbb{E}_{p \sim p_{\text{P}}(p)} [D_p (p)] - \mathbb{E}_{h \sim p_{\text{H}}(h)} [D_p(G_h(h))] \\
\mathcal{L}_{\text{cycle}} &= \mathbb{E}_{p \sim p_{\text{P}}(p)} \left[\|G_h(G_p(p)) - p\|_2^2\right] \nonumber \\
&\quad + \mathbb{E}_{h \sim p_{\text{H}}(h)} \left[\|G_p(G_h(h)) - h\|_2^2\right]
\end{align*}
\vspace{-0.2em}

The total loss combines adversarial and cycle consistency losses, with $\lambda_{\text{cycle}} = 10$ to balance consistency during training \cite{b26}. By integrating Wasserstein loss and cycle consistency loss, we hypothesise that CycleWGAN achieves superior style transfer with greater training stability, producing higher-quality handwritten music images that are well-suited for training OMR systems.

\section{Experiments and Hyperparameter Tuning}
This section outlines the specific hyperparameters and training techniques used for DCGAN, ProGAN, and CycleWGAN models to optimise their performance and address stability issues during training.

\subsection{DCGAN}  

This work implements Label Smoothing for training DCGANs \cite{b28} which is a powerful regularisation technique. The discriminator component of the GAN acts as a classifier, and having hard values like 0.0 and 1.0 for fake and real labels makes the discriminator overconfident and prone to adversarial attacks. Hence, we chose smooth label values 0.1 and 0.9 for fake and real, respectively. All the weights were initialised using a Normal distribution with mean 0.0 and standard deviation 0.02 \cite{b36}.
Both the Generator and Discriminator networks are trained using the Adam optimiser, with a learning rate of 0.0002 , beta values of 0.5 and 0.999 and LeakyRelu in the Discriminator was given with a slope value of 0.2. 
These decisions were made according to the suggestions given by \cite{b36} to gain good training stability. The network accepts images of 64x64 resolution as input and outputs images with the same 64x64 resolution and was trained for 100 epochs as shown in Figure \ref{fig:picture10}. This figure also shows a constant and stable trend in loss for the discriminator. However, while the Generator does show a downward trend in the training loss, it is unstable and begins to display oscillations as it progresses through training.

\subsection{ProGAN}  

The ProGAN model was implemented using tailored hyperparameters and regularisation techniques to optimise performance and stability. A learning rate of 0.001 was set for both the generator and discriminator, ensuring balanced updates to the parameters of the model. The latent dimension $z$ was configured to 256, allowing for the generation of complex features, while the initial number of channels was set to 256, defining the feature map size in the first convolutional layer.
Batch sizes were adjusted according to resolution, starting at 32 and decreasing to 16 for higher resolutions, ensuring computational efficiency. Training at each resolution phase lasted for 30 epochs, sufficient for detailed learning and totalling 180 epochs for six steps. The gradient penalty coefficient (\( \lambda_{GP} \)) was fixed at 10 to enforce smooth gradients, mitigating the dominance of the discriminator. The Adam optimiser was employed with \( \beta_1 = 0.0 \) and \( \beta_2 = 0.99 \), parameters known to stabilise GAN training \cite{b25}.

Regularisation included weight scaling in layers to normalise weights during forward propagation, and Pixel Normalisation (PixelNorm) in the generator to maintain consistent feature magnitudes. Minibatch standard deviation was applied in the discriminator to capture intra-batch variance, improving robustness against mode collapse \cite{b4}.

\begin{figure}[H]
  \centering
  \includegraphics[clip, trim=0.3cm 0.1cm 0.1cm 0.1cm, width=1\linewidth]{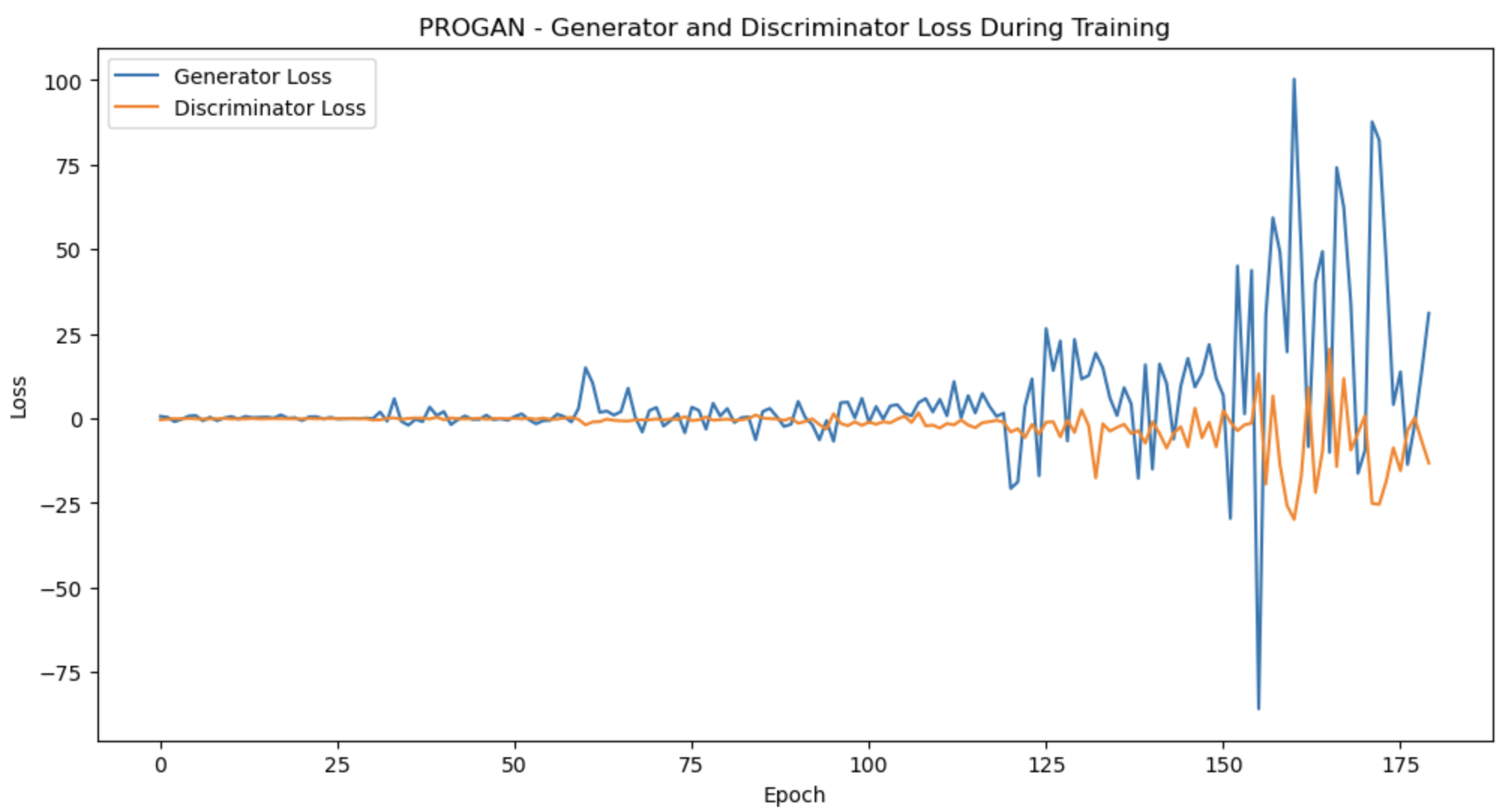}
  \caption{Training loss of ProGAN over 180 epochs demonstrating stable learning progression for lower-resolution images. However, instability appears at higher resolutions, indicating the need for further hyperparameter tuning to enhance stability at advanced stages.}   
  \label{fig:picture_pro}
\end{figure}

From Figure \ref{fig:picture_pro}, we observe a smooth and stable progression of losses during the early training stages for generating lower-resolution images. However, as the number of epochs increases, the curves begin to oscillate, indicating that the model encounters stability issues at higher resolutions. Despite efforts of improving training using WGAN-GP \cite{b22} described earlier, the model faces stability issues only at later stages of the training. Increasing the dimensions of the latent vector $z$ may enable the generator to learn the distribution of higher resolution images better.

\subsection{CycleWGAN}   
\begin{figure}[htbp]
\centering
  \includegraphics[scale=0.8]{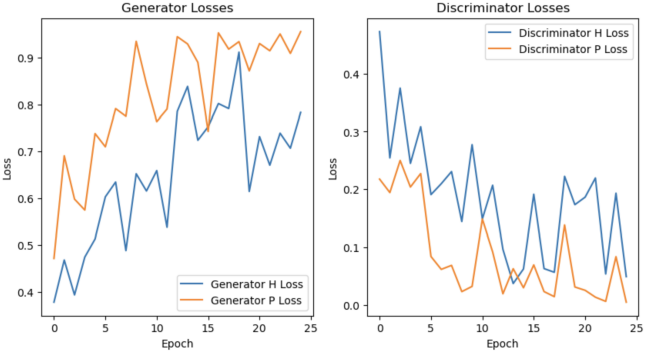}
  \caption{Training loss of (Gen H, Gen P) and (Disc H, Disc P) components of CycleWGAN over 25 epochs.}   
\centering
  \label{fig:picture14}
\end{figure}

To address the limitations of standard GAN models in style transfer tasks, CycleWGAN was implemented with several regularisation techniques and optimised hyperparameters to improve training stability and image quality. These modifications are critical for generating realistic handwritten music images. Regularisation techniques like Instance Normalisation have been used in both the generator and discriminator networks to normalise feature maps. Instance normalisation is known to achieve better results and faster convergence when compared to Batch Normalisation in style transfer tasks \cite{b23}. It is also important to note that Batch Normalisation is incompatible with Wasserstein distance \cite{b22}. The learning rate for both generator and discriminator optimisers is set to \(1 \times 10^{-5}\). The weight for the cycle-consistency loss, \(\lambda\), is set to 10, \cite{b26}.Experimented with custom beta values for Adam optimiser and arrived at (0.5,0.99) \cite{b25}.
The training process spans around 25 epochs, as shown in Figure \ref{fig:picture14}. This figure also indicates a decreasing trend in losses of both the Discriminator H and Discriminator P. As the discriminators become more powerful, Generator H and Generator P tend to struggle, showing a rising trend in their losses. Similar patterns can also be observed in the training trends of the U-Net architecture of \cite{b7}.

\section{Results and Discussion}

This section presents a comprehensive evaluation of the performance of three GAN models, DCGAN, ProGAN, and CycleWGAN, applied to the task of generating handwritten music sheets. The evaluation is divided into two parts: a qualitative analysis, focusing on visual inspection and object detection, and a quantitative analysis, using standard GAN performance metrics. These results help assess the suitability of each model for the practical application of improving OMR systems.


\begin{figure}[htbp]
\centering
\hspace{0.05cm} 
\subfloat[]{%
  \includegraphics[width=.23\linewidth]{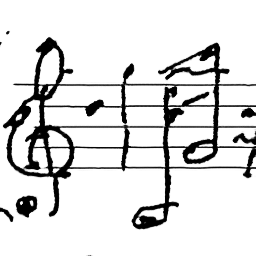}%
  \label{fig:cyc_gan_wgan}%
}
\hspace{0.05cm} 
\subfloat[]{%
  \includegraphics[width=.23\linewidth]{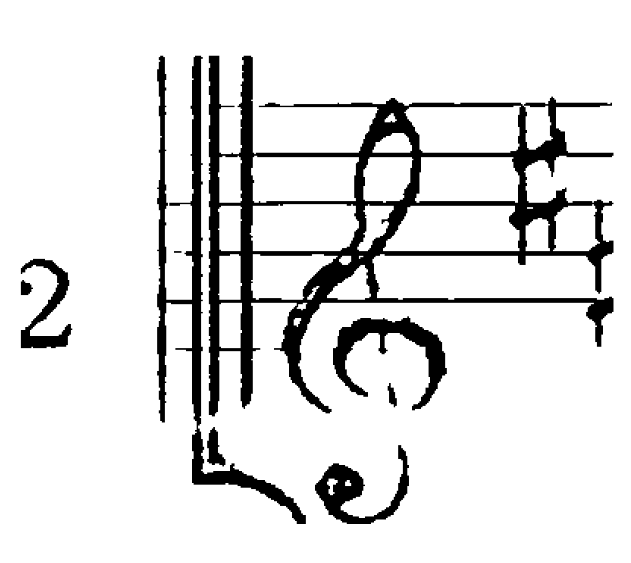}%
  \label{fig:cyc_gan_gan}%
}
\hspace{0.05cm} 
\subfloat[]{%
  \includegraphics[width=.23\linewidth]{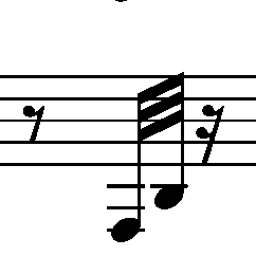}%
  \label{fig:cyc_gan_printed}%
}
\hspace{0.05cm} 
\subfloat[]{%
  \includegraphics[width=.23\linewidth]{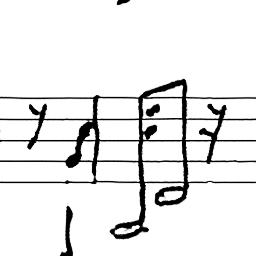}%
  \label{fig:cyc_gan_handwritten}%
}
\caption{(a) CycleWGAN-generated handwritten image. (b) CycleGAN-generated handwritten image. (c) Printed image input. (d) Handwritten image generated by CycleWGAN from the printed image.}
\label{fig:combined_figure}
\end{figure}

\subsection{Qualitative Evaluation} 

The qualitative evaluation aims to visually inspect the generated handwritten music images and assess how well they replicate the intricate details of human handwriting. Additionally, we evaluate the ability of a pre-trained object detection model to recognise and classify musical symbols in the generated images. This analysis provides insights into the visual fidelity of each GAN model and highlights the practical challenges in applying these images to OMR tasks.


\begin{figure}[h]
\centering
\begingroup
\newcommand{\InclGr}[1]{\includegraphics[width=2.3cm,height=2.3cm,valign=t]{#1}}%
\setlength{\tabcolsep}{0.5em} 
\renewcommand{\arraystretch}{1.4} 
\begin{tabular}{*3{c@{\hspace{0.3cm}}}} 

\InclGr{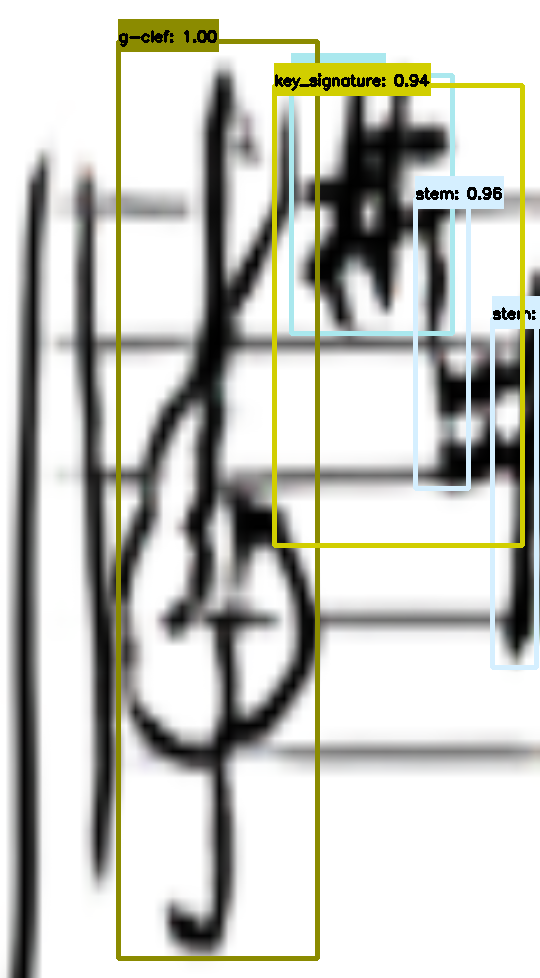} 
& \InclGr{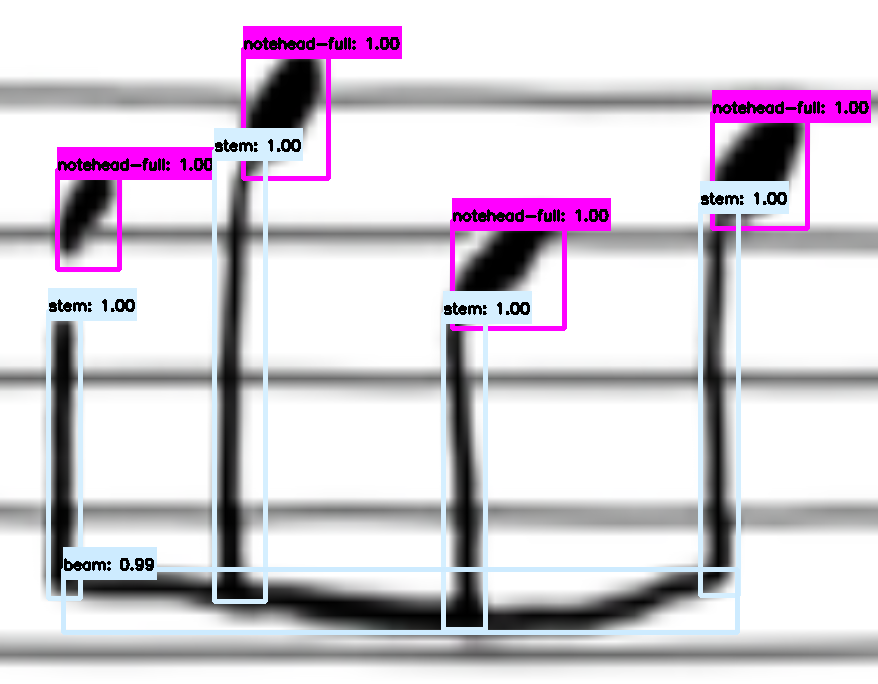} 
& \InclGr{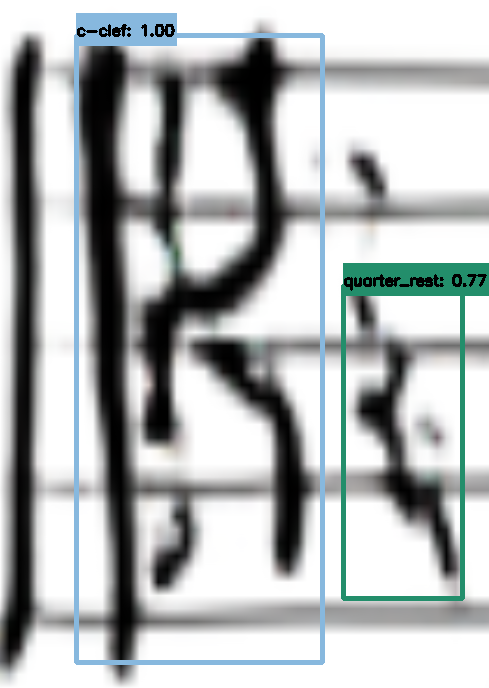}\\
\multicolumn{3}{c}{PROGAN} \\ 

\InclGr{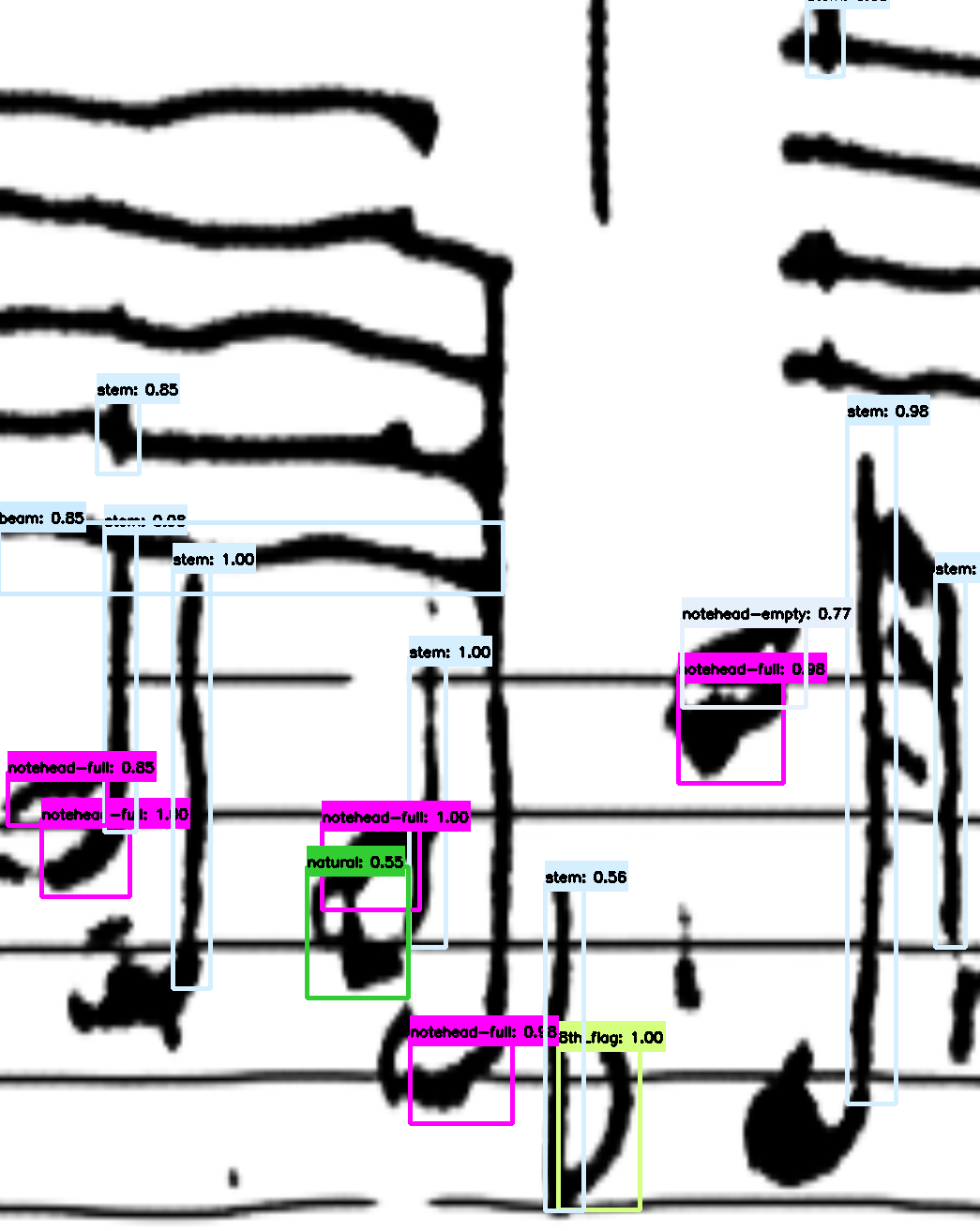} 
& \InclGr{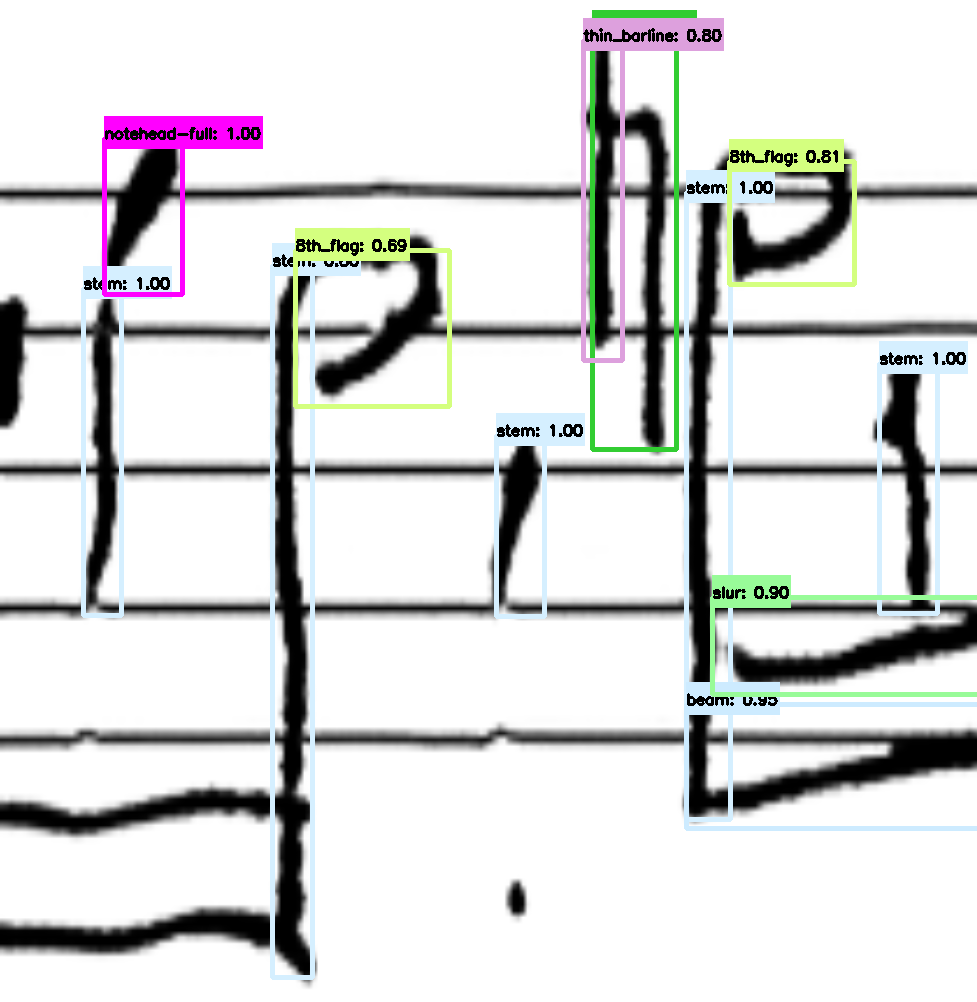} 
& \InclGr{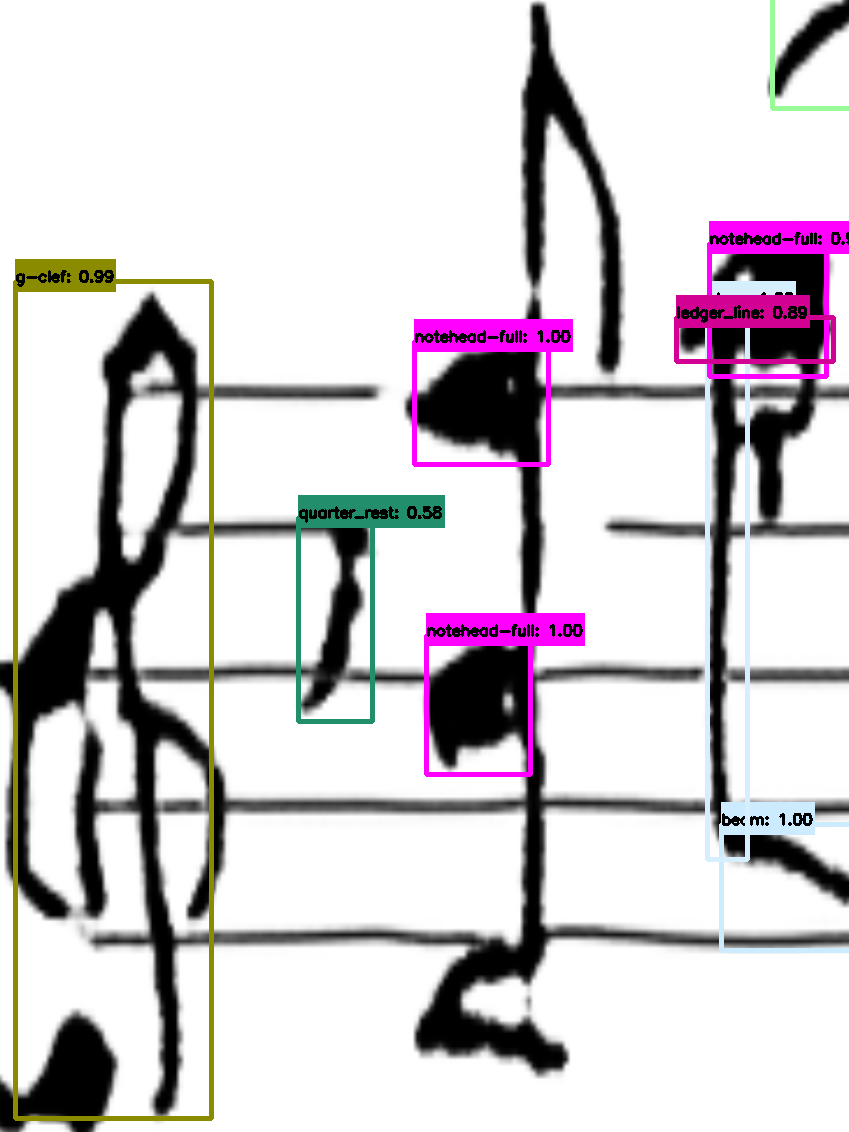}\\
\multicolumn{3}{c}{CycleWGAN} \\ 

\InclGr{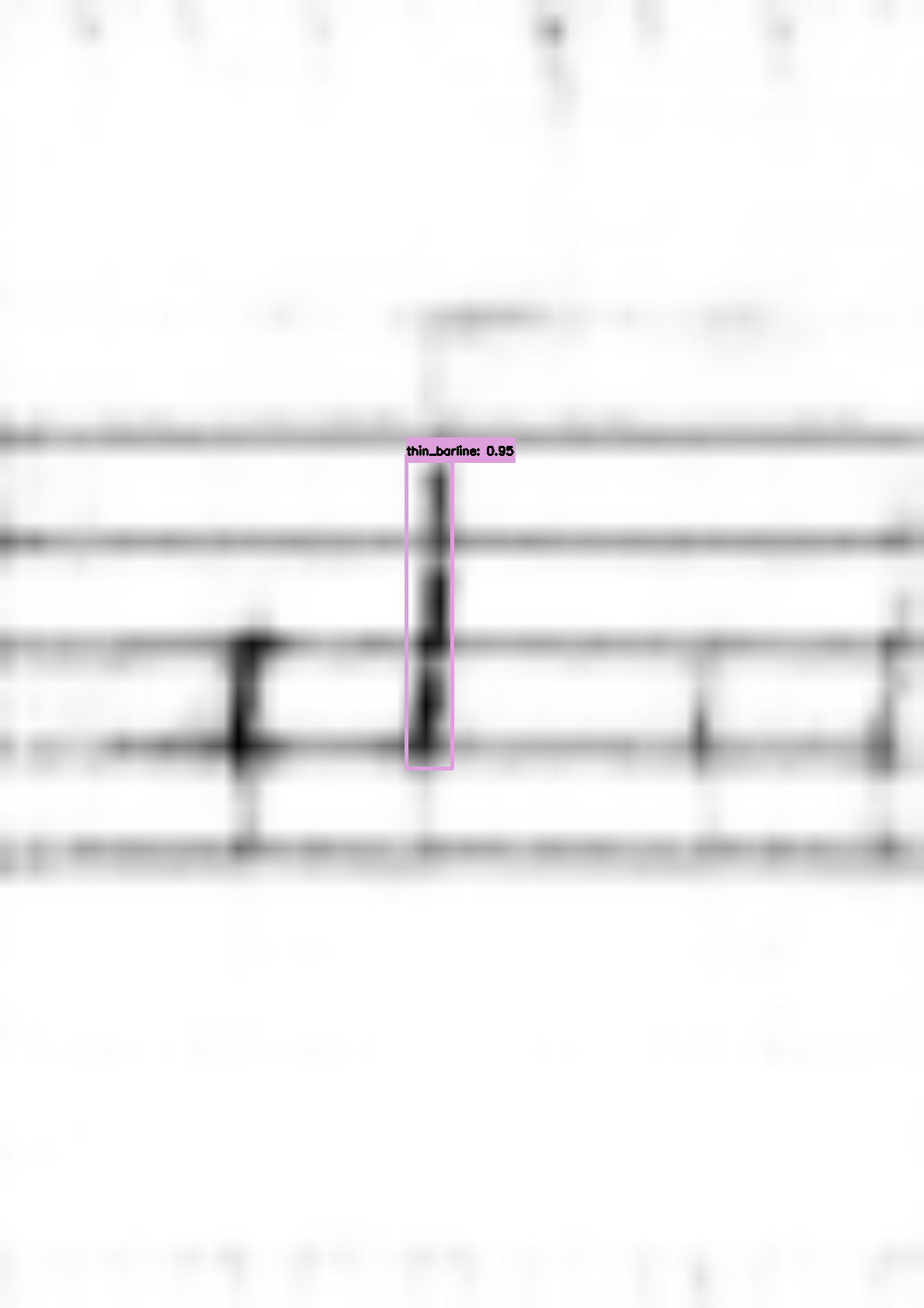} 
& \InclGr{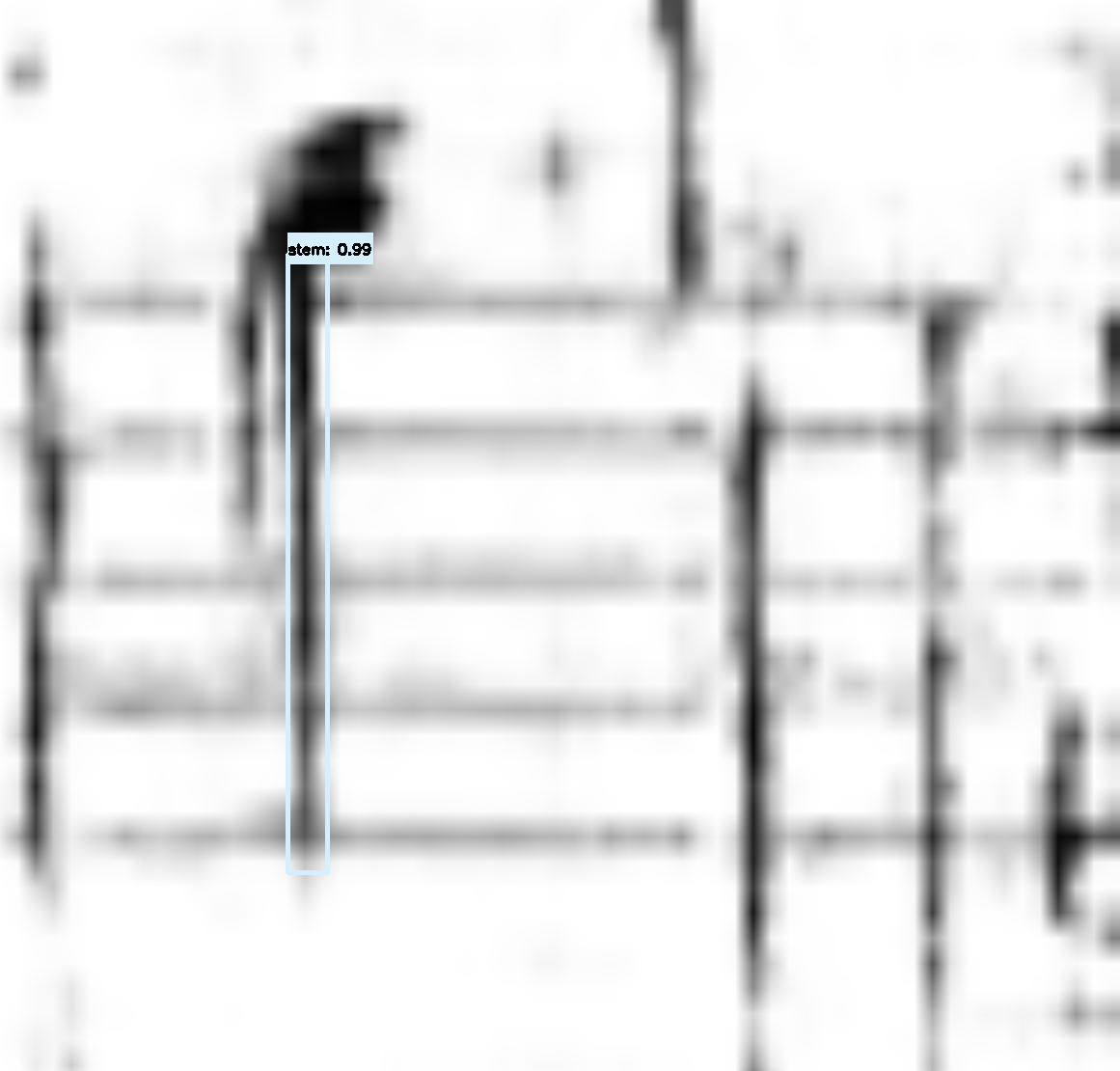} 
& \InclGr{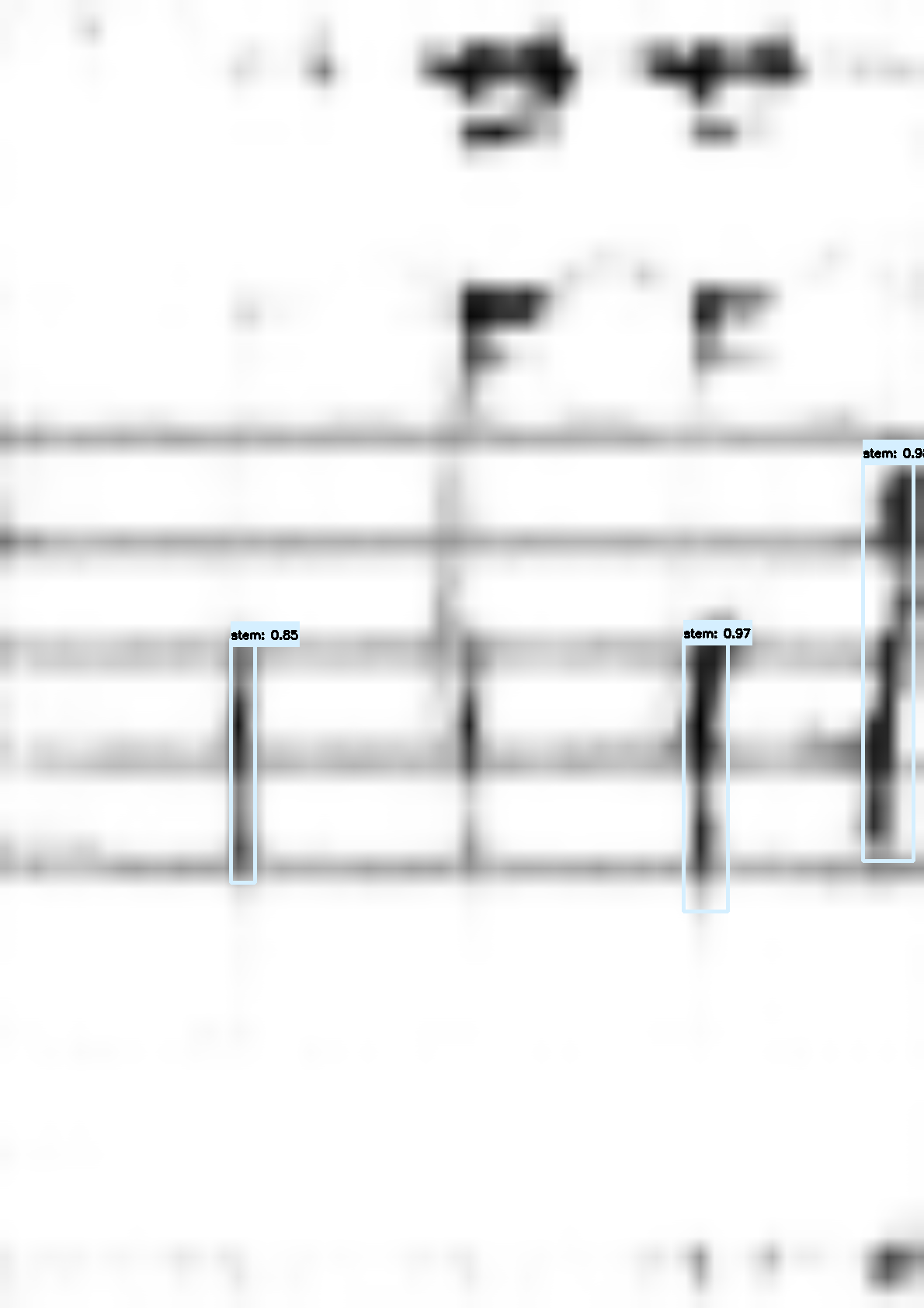}\\
\multicolumn{3}{c}{DCGAN} \\ 

\end{tabular}
\endgroup
\caption{Object detection results on PROGAN, CycleWGAN, and DCGAN generated images.}
\label{fig:obj_detection_results}
\end{figure}




The qualitative analysis of the generated images highlights significant differences in the ability of each GAN model to capture the intricate details of handwritten music. As seen in Figure \ref{fig:obj_detection_results}, all GAN models consistently generate images with five staff lines, a key feature of music notation. However, the performance of these models varies considerably when evaluated for the accuracy of musical symbol generation.

\textbf{DCGAN} produces the least realistic outputs, struggling to capture the complexity of handwritten music distribution due to its limited architecture and the low resolution of its generated images (64x64). Despite regularisation efforts such as label smoothing, the generated images suffer from significant mode collapse, resulting in repetitive and low-quality symbols that fail to resemble handwritten music (see Figure \ref{fig:combined_figure}(c)). This affects the object detection model’s ability to recognise musical symbols effectively, as demonstrated by poor detection scores in Figure \ref{fig:obj_detection_results}.

\textbf{ProGAN}, on the other hand, generates more realistic and detailed images, with symbols such as beamed notes, G-Clefs, and barlines appearing with higher frequency and clarity. However, visual inspection reveals an imbalance in symbol distribution, with common symbols like notes and clefs being overrepresented, while rarer symbols like rests and dynamics are often missing. This imbalance likely stems from biases in the training data and the unconditional nature of the image generation process \cite{b14}. Although ProGAN produces cleaner images than DCGAN, the model still suffers from resolution limitations (128x128), which affect the finer details of complex musical symbols.

In contrast, \textbf{CycleWGAN} demonstrates superior style transfer abilities, generating images that closely resemble human handwriting (Figure \ref{fig:combined_figure}(d)). The images produced by CycleWGAN are not only more stylistically coherent but also better suited for symbol recognition. However, the model introduces gaps and inconsistencies in the transfer of certain musical symbols, such as note heads and accidentals, resulting in intermittent pitch changes and missing elements in the generated images \cite{b7}. These inaccuracies, visible in Figure \ref{fig:cyc_gan_handwritten}, pose challenges for accurate object detection, as certain symbols are not fully represented, leading to incorrect detections by the pre-trained model.

\begin{figure}[htbp]
\centering
\begingroup
\newcommand{\InclGr}[1]{\includegraphics[scale=0.23,valign=t]{#1}}%
\setlength{\tabcolsep}{0.3em} 
\renewcommand{\arraystretch}{1.2} 
\begin{tabular}{cccc} 
    \InclGr{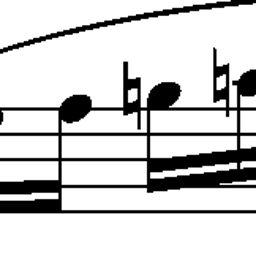} &
    \InclGr{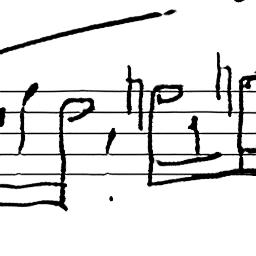} &
    \InclGr{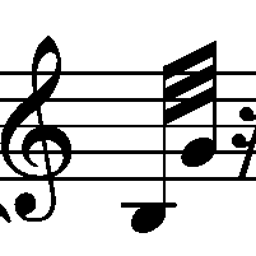} &
    \InclGr{fake_pr_46_2868.png} \\
    
    \InclGr{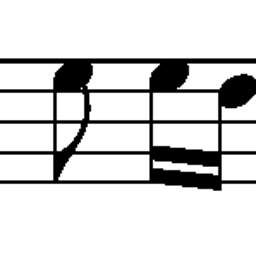} &
    \InclGr{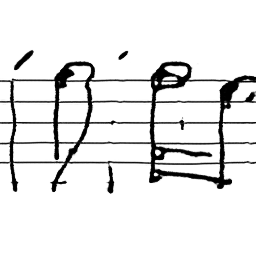} &
    \includegraphics[scale=0.17,valign=t]{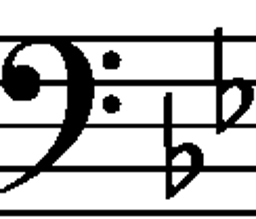} &
    \InclGr{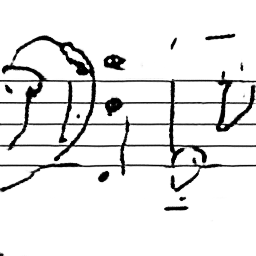} \\
\end{tabular}
\endgroup
\caption{Printed music images and their corresponding images in the handwritten domain translated by CycleWGAN.}
\label{fig:cyc_gan_comparison1}
\end{figure}

\subsection{Quantitative Evaluation}

We evaluate the performance of DCGAN, ProGAN, and CycleWGAN using Inception Score (IS), Frechet Inception Distance (FID), and Kernel Inception Distance (KID). While these metrics are effective for evaluating generative models in other domains, their direct applicability to tasks such as handwritten music synthesis is less established. Therefore, this study supplements these metrics with qualitative evaluations to ensure a comprehensive analysis.

The IS evaluates the quality and diversity of generated images by using a pre-trained classifier, typically an Inception network, to measure how confidently images are assigned to various classes. For handwritten music, this metric verifies if the generated images exhibit sufficient variety and clarity. The IS is defined as:

\begin{equation}
IS = \exp \left( \mathbb{E}_{x \sim p_g} D_{KL} (p(y|x) || p(y)) \right)
\end{equation}

where \( p(y|x) \) is the conditional probability of class \( y \) for image \( x \), and \( p(y) \) is the marginal probability. Higher IS values indicate realistic, diverse outputs. 

The FID measures the similarity between the feature distributions of real and generated images using a pre-trained Inception network. It compares the means (\(\mu_r\), \(\mu_g\)) and covariances (\(\Sigma_r\), \(\Sigma_g\)) of feature representations:

\begin{equation}
FID = \lVert \mu_r - \mu_g \rVert^2 + \text{Tr}(\Sigma_r + \Sigma_g - 2(\Sigma_r \Sigma_g)^{1/2})
\end{equation}

Lower FID values indicate greater similarity to real images. This metric is well-validated in generative model studies for tasks demanding detailed, realistic outputs, such as medical imaging and scene synthesis. 
The KID is an unbiased metric that, like FID, assesses the similarity between real and generated image distributions by estimating the Maximum Mean Discrepancy (MMD) between their feature representations:

\begin{align}
    KID = &\frac{1}{m(m-1)} \sum_{i \neq j} k(x_i, x_j) 
    + \frac{1}{n(n-1)} \sum_{i \neq j} k(y_i, y_j) \nonumber \\
    &- \frac{2}{mn} \sum_{i, j} k(x_i, y_j)
\end{align}
where \( k \) is a polynomial kernel, and \( x \) and \( y \) are features from real and generated images. KID is advantageous for smaller sample sizes due to its unbiased nature, complementing FID with reliable performance on moderate-sized evaluation sets used in this study.

These metrics were selected based on their established use in generative model evaluation. The combination of IS, FID, and KID allows for a comprehensive analysis that addresses both image quality and diversity, aspects critical for synthesising complex handwritten music images. Previous studies [cite key papers] have demonstrated the effectiveness of FID and KID in assessing GAN outputs in fields such as artistic style transfer, medical imaging, and OMR-related applications, ensuring that these metrics are appropriate and validated for our context.

\begin{table}[H]
\centering
\caption{Evaluation on DCGAN, PROGAN, and CycleWGAN}
\setlength{\tabcolsep}{11pt} 
\renewcommand{\arraystretch}{1.3} 
\begin{tabular}{|c|c|c|c|}
\hline
\textbf{GAN Model}    & \textbf{IS}               & \textbf{FID}   & \textbf{KID}    \\ 
\hline
\text{DCGAN}          & 2.17483 $\pm$ 0.14473     & 210.178        & 0.21266         \\ 
\hline
\text{PROGAN}         & 2.00708 $\pm$ 0.19896     & 49.54          & 0.16689         \\ 
\hline
\text{CycleWGAN}      & 2.29289 $\pm$ 0.19285     & 41.87          & 0.05            \\ 
\hline
\end{tabular}
\label{result_table}
\end{table}
\textbf{CycleWGAN} achieves the best performance across all metrics, with an FID of 41.87 and a KID of 0.05, indicating the highest image quality and diversity among the three models (Table \ref{result_table}). However, the relatively high (IS of 2.29 suggests that while the images are visually coherent, there may still be issues with mode collapse, as the model tends to generate similar images that lack sufficient diversity \cite{b30}.
  
\textbf{ProGAN}, although producing clearer images than DCGAN, performs worse than CycleWGAN in terms of FID (49.54) and KID (0.17), reflecting lower image diversity and quality. The results align with the qualitative findings, where ProGAN generated visually accurate images but struggled with maintaining symbol diversity due to lower resolution and imbalanced training data.

\textbf{DCGAN} performs the worst, with an FID of 210.18 and KID of 0.21, highlighting its inability to generate realistic handwritten music images. These metrics indicate significant issues with both image quality and diversity, consistent with the observed mode collapse during qualitative analysis.

\subsection{Principal Component Analysis}

\begin{figure}[htbp]
  \centering

  \subfloat[]{%
    \includegraphics[clip, trim=1.5cm 1.2cm 2cm 1.8cm, width=0.9\linewidth]{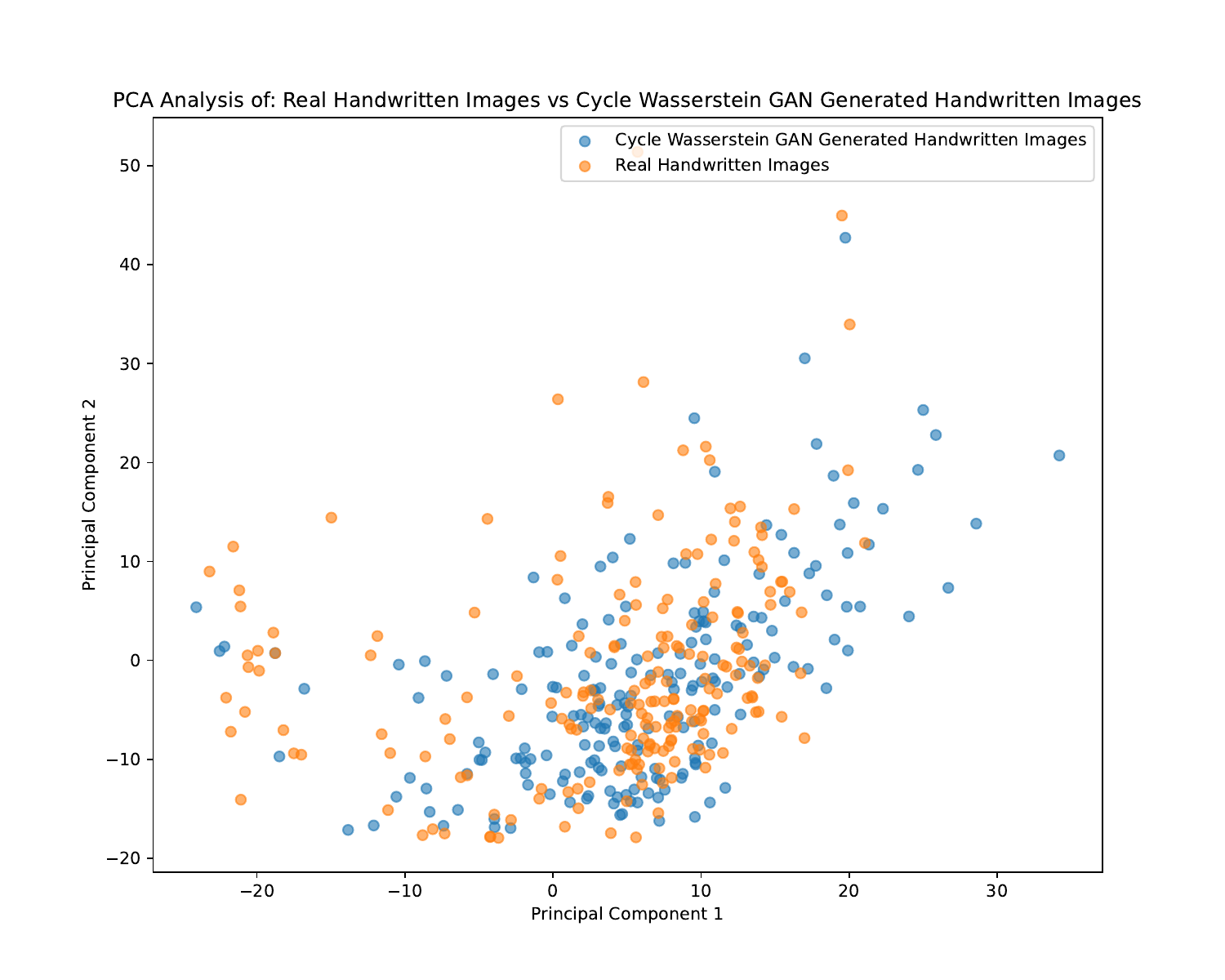}%
    \label{fig:cycle_wgan_pca}%
  }
  \vspace{0.02cm}  
  
  \subfloat[]{%
    \includegraphics[clip, trim=1.5cm 1.2cm 2cm 1.8cm, width=0.9\linewidth]{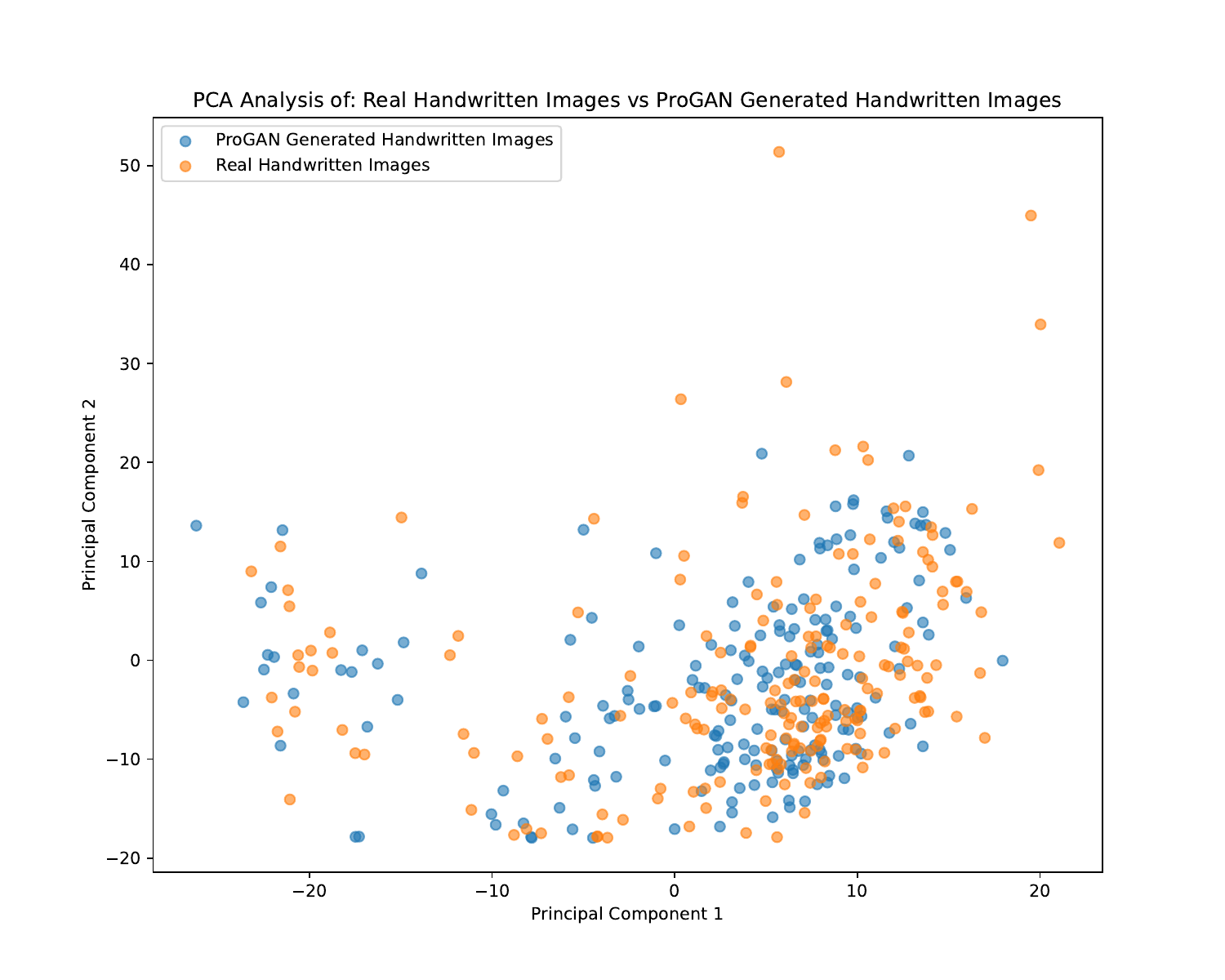}%
    \label{fig:progan_pca}%
  }
  \vspace{0.02cm}  

  \subfloat[]{%
    \includegraphics[clip, trim=1.5cm 1.2cm 2cm 1.8cm, width=0.9\linewidth]{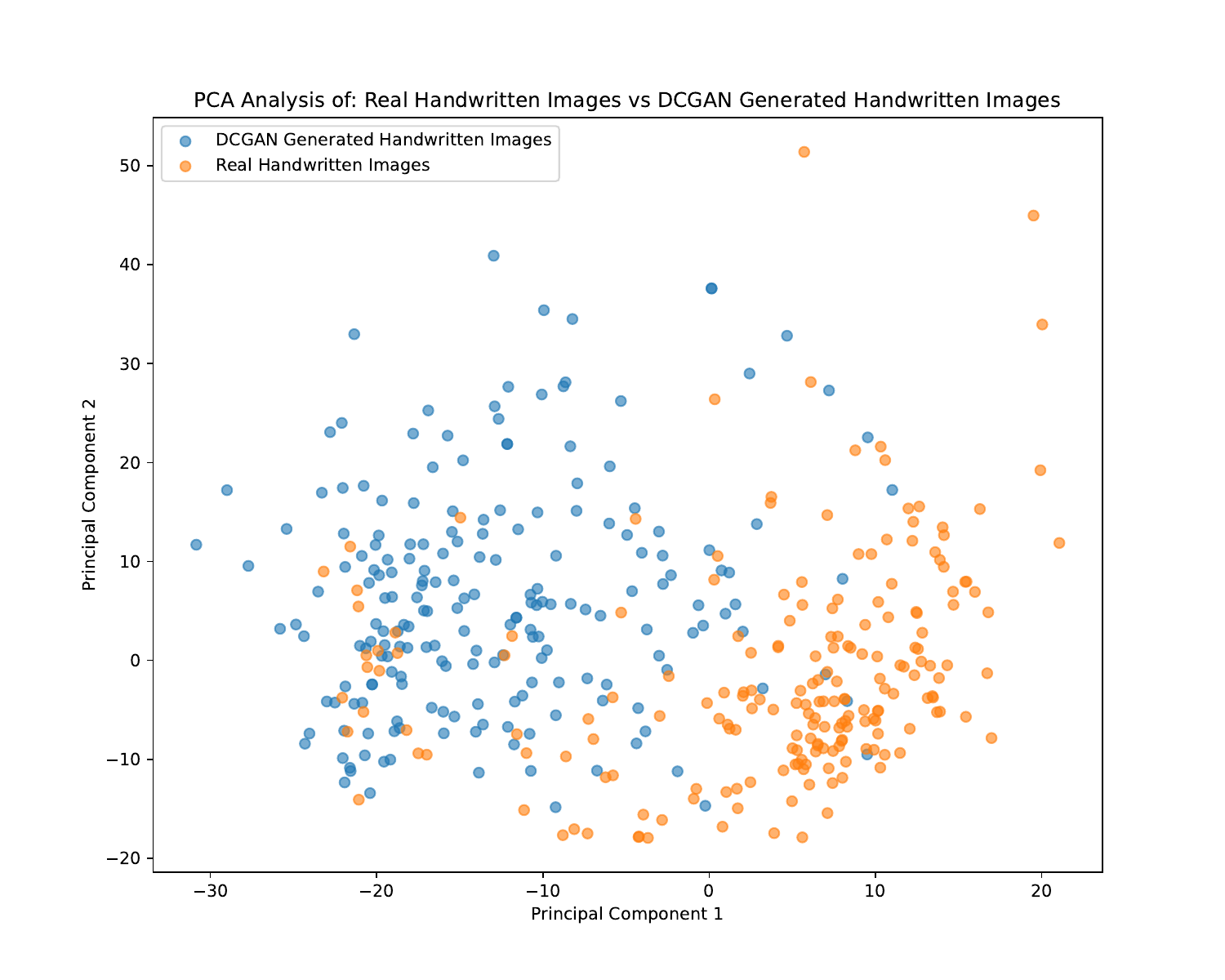}%
    \label{fig:dcgan_pca}%
  }
  \caption{PCA analysis of test data for individual GAN models comparing real and generated handwritten images: (a) CycleWGAN, which includes real handwritten and generated images; (b) ProGAN, displaying real handwritten and generated images without printed features; (c) DCGAN, showing real handwritten and generated images.}
  \label{fig:pca_combined}
\end{figure}

To further analyse the performance of the GAN models, Principal Component Analysis (PCA) was used to visualise the similarity between real and generated handwritten images for each model in Figure \ref{fig:pca_combined}. Figure \ref{fig:cycle_wgan_pca} shows that the images generated by CycleWGAN are clustered closest to the real handwritten images compared to Figure \ref{fig:dcgan_pca} and \ref{fig:progan_pca}, demonstrating the model’s success in capturing stylistic features of handwritten music. Printed images are not included for ProGAN and DCGAN as these models do not incorporate printed features in their training.



A deeper analysis, shown in Figure \ref{fig:pca_combined}(b), compares the real handwritten images, real printed images, and the CycleWGAN-generated images. The handwritten images generated by CycleWGAN are clearly grouped closer to the real handwritten images than to the printed ones, indicating effective style transfer from printed to handwritten domains. This result is consistent with the superior performance of CycleWGAN in the FID and KID metrics.

\subsection{Limitations}

While CycleWGAN outperformed other models in generating realistic handwritten music images, several challenges remain. Issues such as gaps in note head transfer and missing musical symbols suggest the need for further refinements, such as employing inpainting techniques to correct incomplete style transfers \cite{b7}.

Another limitation is the relatively small evaluation dataset (200 images), which may affect the reliability of metrics like FID, typically requiring larger sample sizes for accurate estimates \cite{b30}. Future work should focus on increasing the dataset size and exploring alternative evaluation metrics, such as KID, which are better suited to smaller datasets.

\section{Conclusion and Future Work}

This research builds upon existing work that addresses the complex challenge of synthesising handwritten music sheet data. By exploring advanced deep learning and training techniques, specifically with GANs, we have contributed to the development of more realistic and diverse datasets for OMR systems. Our comparative analysis of multiple GAN models provides a robust evaluation framework, highlighting the strengths and limitations of each model in synthesising handwritten music.


While DCGAN struggled to accurately capture the distribution of handwritten music images, both CycleWGAN and ProGAN demonstrated more promising results. Our proposed CycleWGAN outperformed the earlier CycleGAN model proposed by \cite{b7}, achieving a lower FID score, a higher IS, and improved style transfer, as shown in Figure \ref{fig:cyc_gan_comparison1}. These improvements are a step forward in addressing the scarcity of training data for handwritten music, offering a more stable and effective model for image translation.
However, CycleWGAN’s translation process still introduced gaps and inaccuracies, particularly in the imperfect transfer of musical noteheads. These flaws, along with random pitch shifts, render the generated images unsuitable for direct use in training OMR systems. By contrast, ProGAN produced more complete and balanced images, although it exhibited an imbalance in the frequency of certain musical symbols, such as notes appearing more frequently than rests or dynamic markings.

Future work could focus on addressing these limitations and further improving the resolution and quality of generated images. Incorporating Wasserstein GAN with Gradient Penalty (WGAN-GP) \cite{b22}, which was used successfully in ProGAN training, could enhance the training stability and quality of CycleWGAN outputs. Furthermore, expanding the use of inpainting techniques, as discussed by \cite{b7} or exploring new generative models, such as diffusion models, may lead to even more accurate and diverse outputs. These advances could contribute to more effective OMR systems and create opportunities for artistic and editorial applications of synthesised handwritten music styles.

Increasing the resolution of ProGAN-generated images to 1024x1024 and expanding the latent vector dimension to 512 could further enhance both the quality and diversity of the generated datasets \cite{b4}. These improvements would likely lead to more realistic and high-resolution handwritten music images suitable for a wider range of OMR applications.

Another exciting area of exploration is the application of diffusion models \cite{b21}, which are gaining attention in the field of deep generative models. While diffusion models offer new opportunities for generating highly detailed images, there remains a gap in their ability to produce non-square images, a crucial requirement for full sheet music generation. The ultimate goal is to generate entire handwritten music sheets, complete with staff lines and musical symbols, in a style that closely resembles human handwriting.

This study demonstrates the potential for developing a dataset of annotated printed and handwritten music image pairs. This contribution could significantly improve the training of OMR systems by providing access to larger and more diverse datasets.

By addressing these future directions, we aim to overcome the limitations of current models and move closer to the goal of fully automated and accurate handwritten music recognition systems.

\section*{\uppercase{Acknowledgements}}

The authors acknowledge the support of the AI and Music CDT, funded by UKRI and EPSRC under grant agreement no. EP/S022694/1, and our industry partner Steinberg Media Technologies GmbH for their continuous support.

\end{document}